\documentclass[letterpaper, 10 pt, journal, twoside]{ieeetran}  

\usepackage{amsmath} 
\usepackage{amssymb}  
\usepackage{graphicx} 
\usepackage{bm}
\usepackage{url}

\newcommand{\argmax}{\mathop{\rm arg~max}\limits}

\title{
A Tendon-driven Robot Gripper with Passively \\ Switchable Underactuated Surface and its \\Physics Simulation Based Parameter Optimization
}

\author{Tianyi Ko$^{1}$%
\thanks{Manuscript received: Feb, 19, 2020; Revised May, 15, 2020; Accepted Jun, 11, 2020.}
\thanks{This paper was recommended for publication by Editor Hong Liu upon evaluation of the Associate Editor and Reviewers' comments.} 
\thanks{$^{1}$T. Ko is with Preferred Networks Inc., 3F Otemachi-building, 1-6-1 Otemachi, Chiyoda-Ku, Tokyo, Japan.
    {\tt\footnotesize tko@preferred.jp}
}
\thanks{Digital Object Identifier (DOI): see top of this page.}
}

\begin{document}

\maketitle

\markboth{IEEE Robotics and Automation Letters. Preprint Version. June, 2020}
{Tianyi Ko: Optimization of a Gripper with Underactuated Surface} 


\begin{abstract}
In this paper, we propose a single-actuator gripper that can lift thin objects lying on
a flat surface, in addition to the ability as a standard parallel gripper.
The key is a crawler on the fingertip,
which is underactuated together with other finger joints and 
switched with a passive and spring-loaded mechanism.
While the idea of crawling finger is not a new one,
this paper contributes to realize the crawling without additional motor.
The gripper can passively change the mode from the parallel approach mode
to the pull-in mode, then finally to the power grasp mode, according to the grasping state.
To optimize the highly underactuated system,
we take a combination of black-box optimization and physics simulation of the whole grasp process.
We show that this simulation-based approach can effectively consider
the pre-contact motion, in-hand manipulation, power grasp stability, and even failure mode,
which is difficult for the static-equilibrium-analysis-based approaches.
In the last part of the paper, we demonstrate that a prototype gripper with the proposed structure
and design parameters optimized under the proposed process successfully power-grasped a thin sheet,
a softcover book, and a cylinder lying on a flat surface.
\end{abstract}


\section{Introduction}

\IEEEPARstart{O}{ne} of the major applications of robots is manipulation, and robot hands play a critical role in such contexts.
A large number of robot hands already exceed the academic research stage and are commercially available.
In addition to the rigid parallel grippers~\cite{onrobot,robotiq_parallel},
underactuated ones~\cite{robotiq_envelop, robotis} are also available, which can adapt a variety of object shapes
with a single actuator.
To handle soft objects such as food, pneumatically-driven soft grippers~\cite{softrobotics,piab} 
have emerged in the market.
For tasks that require grasp-mode switch,
three-fingered hands with multiple actuators~\cite{robotiq_3finger, barrett,righthand}
are the option.
If further in-hand manipulation~\cite{openai_inhand} is required,
high degree-of-freedom and degree-of-actuation anthropomorphic hands~\cite{shadow}
are the solution.

Despite their high versatility, high degree-of-actuation hands are still limited for research purposes,
and in most applications single-actuator grippers are adopted due to their low cost, high mechanical reliability, and easiness of control.
Thus, improving the versatility of simple single-actuator grippers can straightforwardly contribute to the real application of robots.
A weak point of simple grippers is that they have difficulty in grasping thin objects lying on a flat surface,
which is a frequently required task for robots outside well-designed factory environments, e.g., 
in the household or jig-less manufacturing scene.
The surface prevents the fingers from enveloping the object while the precision grasp with the fingertip is sensitive
against object deformation or misalignment.

\begin{figure}[t]
    \begin{center}
        \includegraphics[width=0.55\columnwidth, bb=35 35 205 190, clip]{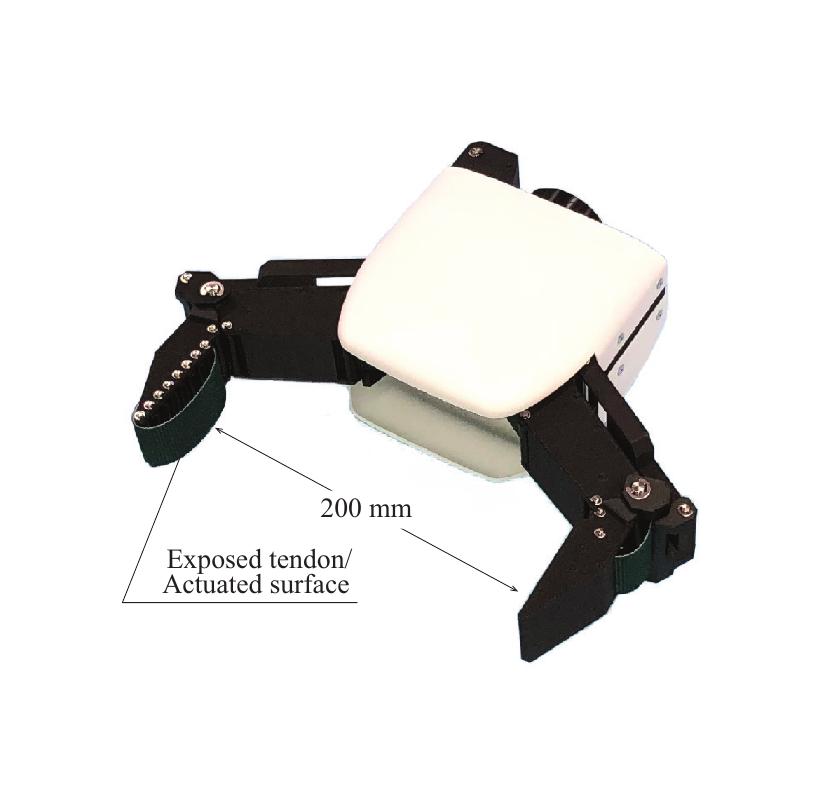}
        \caption{
            Outlook of the developed gripper.
            The essential component is a crawler on the fingertip,
            which is underactuated together with other finger joints and switched with a passive mechanism.
        }
        \label{fig:hand_outlook}
    \end{center}
\end{figure}

In the case of the human hand, we have two choices of motion to lift thin objects from flat surfaces depending on the condition.
One is to use the nail to insert the finger under the object.
As the other choice, we can use a cyclic conveying motion of the index and middle finger to pull the object inside the hand.
To mimic the former motion, Babin et al.~\cite{babin2019stable,babin2018picking} proposed to use a passive joint and sharp-tip phalanx
to insert the finger under the object.
To maintain the contact, force control~\cite{babin2019stable} or a passive compliance~\cite{babin2018picking} was introduced.
The latter motion can be achieved by adding a conveyer or crawler
on the surface of the fingers~\cite{velvet_fingers, yale_belt_hand}.
In this case, the control is straightforward but additional actuators are needed to drive the crawler.
In this paper, we focus on the latter case and propose a passive switching mechanism to underactuate both the joints
and crawler, requiring no additional actuator.

Given the basic structure of a hand, deciding the design parameters such as the geometric property of the links
and the actuation parameters of the joints is also a challenging problem.
While there are a variety of successful works on the optimization of 
underactuated grippers~\cite{ciocarlie2010data, velo_gripper, dong2018geometric, chen2019tendon},
most works consider the static equilibrium of the grasp wrench.
The exception is the work by Chen et al.~\cite{chen2019tendon}, considering both pre-contact posture synergy
and post-contact torque synergy.
Grasping with the actuated surface, though, is an in-hand manipulation process,
which makes analytical formulation of the contact state transition difficult.
Another difficulty with the static equilibrium analysis is that it cannot evaluate the failure mode,
i.e., what happens after a slip occurs due to disturbance or object deformation.
When pinching at the very tip of the finger, the grasp is not {\it reliable} because
a small slip will directly result in a drop.
With the same static equilibrium, pinching in the middle of the finger is more {\it reliable} due to the margin.
Another failure mode that static analysis can not handle
is the case where a disturbance and slip leads the object to fall into a globally more stable grasp configuration.
In this paper, we employ a straightforward but powerful approach,
which is to simulate the whole grasp process, i.e.,  pre-contact approaching, in-hand manipulation, and power grasp
until the hand finally drops the object, and employ a general-purpose black-box optimization tool to solve this problem.

The contribution of this paper is (i) proposal of a new gripper structure
with a crawling surface to lift thin objects but without additional actuators to drive it,
(ii) proposal of a hand optimization framework that can consider all of the pre-contact motion, in-hand manipulation, power grasp stability, and failure mode, and
(iii) experimental evaluation of the proposed methods, including the required control scheme. 
The paper is organized as follows.
Section~\ref{sec:related_works} summarizes the related works.
We explain the idea of the proposed mechanism in section~\ref{sec:mechanism}.
Section~\ref{sec:optimization} details the modeling of the system and its optimization.
Section~\ref{sec:evaluation} describes the evaluation of the prototyped gripper.
We have a short discussion in section~\ref{sec:discussion} and conclude the paper in section ~\ref{sec:conclusion}.

\section{Related Works}
\label{sec:related_works}
\subsection{Grippers with Actuated Surface}
The essential component of the proposed gripper is a crawler on the fingertip,
which is underactuated together with other finger joints and switched with a passive mechanism.
The idea of an actuated finger surface itself is not a new one.
Tincani et al.~\cite{velvet_fingers} proposed a two-fingered gripper with crawlers on all of the four phalanxes.
The independently-actuated crawlers enabled a high in-hand manipulation performance.
Ma and Dollar~\cite{yale_belt_hand} showed that a fixed finger with a crawler improved the performance
to lift objects lying on a flat surface.

In those works, the crawlers require additional motors other than the ones for the joints.
This results in a complex structure and high cost.
It is also not clear that which has more contribution to the grasping performance: 
to use the additional motor to drive a crawler,
or to add another degree of freedom to the fingers.
Kakogawa et al.~\cite{ritsumei_gripper} proposed a gripper whose fingertip crawlers are driven
by the same actuator with the finger joints.
A differential gear unit split the motor torque to both the joints and crawlers.
A braking system on the crawler adds enough friction to avoid crawlers' motion
before the contact.
A difficulty is that the breaking force of the crawler constrains the upper bound of the joint torque:
the motor torque exceeding the braking force is transferred to the crawler;
thus the joint torque cannot exceed the one corresponding to the braking force.

\subsection{Optimal Mechanical Design}
The proposed gripper is highly underactuated with a large number of design parameters,
which are not straightforward to decide.
To optimize them,
the problems are the way to model the contact, to evaluate the grasp quality, and to optimize the score.
In the work by Ciocarlie et al.~\cite{velo_gripper}, the contact is assumed to happen on the center of each phalanx.
A grasping quality is highly evaluated if the reaction force at each contact point is close to a heuristically predefined one.
They proposed a combination of random sampling and gradient descent with numerical deviation to solve the optimization problem.
In the work by Ciocarlie and Allen~\cite{ciocarlie2010data}, 
they acquired the contact points with multiple household items' 3D mesh from {\it GraspIt!} simulator with a force-closure constraint.
They formulated the optimization problem as a Quadratic Program (QP) to minimize grasping wrench
under the constraint of underactuation.
In this case, only the joint driving parameters are optimized and the links' geometric properties are set as constant,
leaving the contact points invariant through the optimization.
Dong et al.~\cite{dong2018geometric} used the genetic algorithm to optimize the geometric properties and driving parameters
of a tendon driven underactuated gripper.
The target of the optimization was to maximize the grasping force, uniformly distribute the contact forces,
and maximize the force transmission efficiency.
They selected primitive elliptical and rectangle objects as the target objects.
Similarly to \cite{velo_gripper}, they sampled multiple joint configurations, but the contact points
were also optimized by adjusting the link thickness.
In the work by Chen et al.~\cite{chen2019tendon},
the optimization problem was to shape the Mechanically Realizable Manifold to the desired synergy.
They used multi-staged dual-layer optimization, which employs QP to solve the inner layer and CMA-ES~\cite{cma-es} for the outer layer.
They treated the geometric feature of the hand as given and optimized the actuation variables.
Similar to \cite{ciocarlie2010data}, they acquired the desired synergy from the {\it GraspIt!} simulator with multiple target objects.

In those works,
the contact points between the hand and object are treated as given and fixed.
For the proposed gripper, however, the fingertip crawler actively changes the relative position of the fingers and object
during the grasping process to end up with a {\it reliable} grasp.
To evaluate the grasp performance of the proposed gripper, we create a physics simulation environment
to simulate the transition of the grasp and apply a simulated disturbance until the object falls.

\section{Passively Switchable Underactuated Surface}
\label{sec:mechanism}

\begin{figure}[t]
    \begin{center}
        \includegraphics[width=0.77\columnwidth]{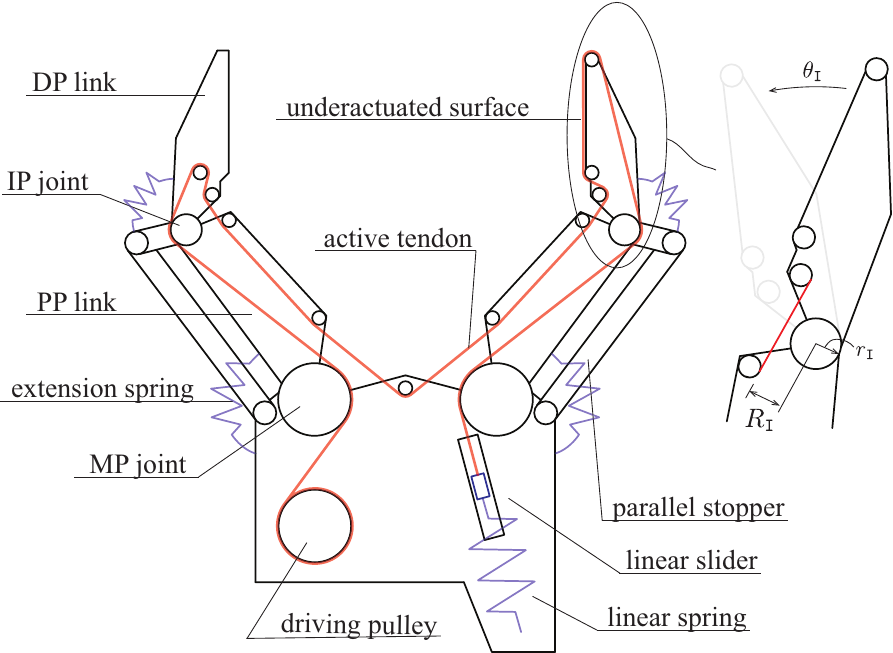}
        \caption{
            Schematic of the proposed robot gripper with an underactuated surface.
            It differs from conventional designs in that the active tendon is exposed to also serve as
            a fingertip crawler.
            An elastic element and linear slider terminates the other end of the tendon from the actuator.
            The range of motion of the crawler corresponds to that of the slider.
        }
        \label{fig:hand_schematic}
    \end{center}
\end{figure}

Figure~\ref{fig:hand_schematic} illustrates the mechanism of the proposed gripper.
The basic structure follows the conventional ones: two opposing fingers with two phalanxes for each finger,
underactuated by a single tendon.
A stopper based on the parallel link mechanism constraints the extension of the IP joint to maintain
the DP links in parallel when the hand closes.
The stopper does not prohibit flexion, allowing enveloping grasp.

\begin{figure}[t]
    \begin{center}
        \includegraphics[width=0.75\columnwidth]{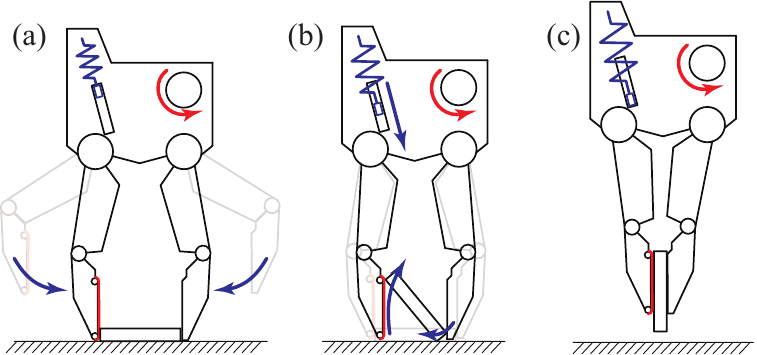}
        \caption{
            Schematics of the passive switching mechanism.
            (a) Before the fingers have contact with the target object, the slider
            does not move due to the pretension of the elastic element.
            (b) After the contact and as tendon tension increases, the deformation of the elastic element
            allows the slider to move, leading a motion of fingertip crawler to pull the object
            inside the hand.
            (c) After the slider reaches the limit, the gripper again serves as a standard gripper.
            This limit of the slider is essential to avoid losing grasping force when
            the friction between the object and crawler is low.
        }
        \label{fig:motion_schematic}
    \end{center}
\end{figure}

The key difference of the proposed structure is that
the active tendon is exposed to also serve as a fingertip crawler.
An elastic element and linear slider terminates the other end of the tendon from the actuator.
The range of motion of the crawler corresponds to that of the termination part
(hereafter, we denote the termination part as "sliding part").
The grasping process consists of three modes.
In the first mode, the extended hand closes to reach the object.
The elastic element on the sliding part has higher stiffness than the extension spring of the finger joint,
resulting in no motion of the slider during this mode.
In this mode, the approaching motion is the same as a parallel gripper without a crawler.
In the second mode, the fingers have contact with the object and the fingers' flexion is interrupted.
As the motor torque increase and tendon tension exceeds the pretension of the sliding part,
the slider starts to move.
This motion leads the exposed tendon on the fingertip to move toward the palm side, pulling the object
inside the gripper.
In the third mode, the slider reaches a mechanical end.
At this stage, the gripper serves as an underactuated gripper without a crawler, 
fully applying the motor torque to the joints.
Figure~\ref{fig:motion_schematic} illustrates the transition of the three modes of grasping.

As illustrated in Fig.~\ref{fig:motion_schematic}, the crawlered finger needs to contact
the object from the side, which means that a non-crawlered conventional hand
should also be able to pick it.
The difference, though, is the robustness against deformation and disturbance.
When a thin object is pinched at the very tip of the fingers,
a small deformation or bending of the object leads the contact wrench to direct to the ejection direction.
The finger also needs to be carefully designed to have enough stiffness.
In addition, the grasp force is limited to avoid buckling.

A major failure mode of grasping is the slip due to impulsive disturbance,
such as acceleration or vibration of the arm and collision with the environment.
Since fingertip pinching has no positional margin, a small slip can cause the object to 
be totally dropped.
In the case of human, we rarely keep the initial pinching configuration through the whole
manipulation; we usually use an in-hand manipulation to switch to power-grasp mode
as soon as the object is lifted from the hard surface.
In the proposed hand, the second mode (b) is equivalent to the in-hand manipulation
from the initial fingertip pinching to a tighter power-grasp.

\section{Simulation-based Design Parameter Optimization}
\label{sec:optimization}
\subsection{Hand Modeling}

\begin{figure}[t]
    \begin{center}
        \includegraphics[width=0.6\columnwidth]{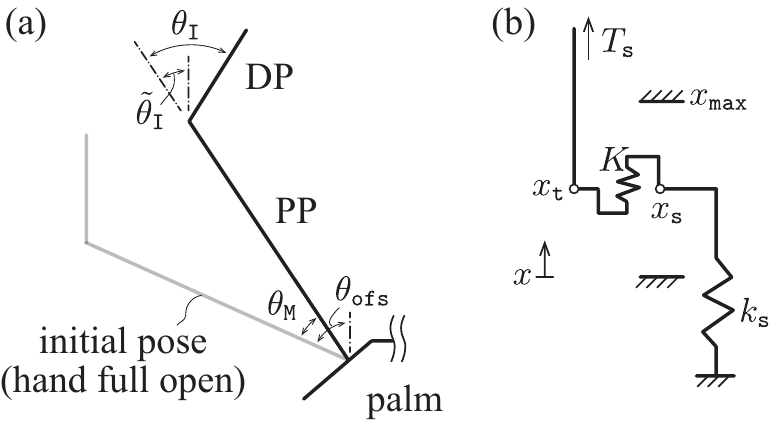}
        \caption{
            Definition of the joint angles (a) and schematic of the model of the
            tendon termination part with slider.
        }
        \label{fig:termination}
    \end{center}
\end{figure}

To decide the design parameters of the gripper, we ran a black-box optimization
with the resulted score of grasp simulations as the target function.
We selected the design parameters $\bm{\theta}$ as follows:
\begin{equation}
    \bm{\theta} = 
        \begin{bmatrix}
        l_{\tt D} &
        l_{\tt P} &
        l_{\tt M} &
        r_{\tt I} &
        R_{\tt I} &
        R_{\tt M} &
        k_{\tt s} &
        T_{\tt pt}
        \end{bmatrix}^T
\end{equation}
Here, $l_{\tt D}, l_{\tt P}$ are the length of the DP and PP links, respectively.
The distance between the two MP joints is represented as $l_{\tt M}$.
As shown in Fig.~\ref{fig:hand_schematic}, the tendon passes each joint twice.
On one side the tendon wraps around a pulley with a radius $r$.
On the other side the tendon directly goes from one link to another.
The moment arm, which is the distance between the joint center and the tendon, varies according to the joint angle.
We use $R$ to express the moment arm when the finger is fully extended.
To avoid collision between the tendon and pulley, $R > r + \delta$ stands,
where $\delta$ represents a margin considering the tendon thickness.
In this work we set $\delta$ as 1 mm.
Those parameters for the IP joint is expressed as $r_{\tt I}$ and $R_{\tt I}$.
For the MP joint, we only set $R_{\tt M}$ as a design variable and assumed $r_{\tt M} = R_{\tt M} - \delta$.
These parameters were treated as the same for the two fingers.
The spring coefficient of the linear spring on the sliding part is expressed as $k_{\tt s}$.
$T_{\tt pt}$ represents the pretension of the spring.

In this work, modeling of the contact point, contact force, grasp wrench, and their transition
are handled by the physics simulator, thus the only part we need to explicitly model
is the joint transmission.
The relationship between the tendon tension and joint actuation torque is expressed as follows:
\begin{eqnarray}
    \begin{bmatrix}
        \tilde{\tau}_{\tt M 1} \\
        \tilde{\tau}_{\tt I 1} \\
        \tilde{\tau}_{\tt M 2} \\
        \tilde{\tau}_{\tt I 2} \\
        \tau_{\tt c} \\
    \end{bmatrix} = 
    \begin{bmatrix}
        R_{\tt M}\sqrt{1+sin(\theta_{\tt M 1})} + r_{\tt M} \\
        R_{\tt I}\sqrt{1+sin(\theta_{\tt I 1})} - r_{\tt I} \\
        R_{\tt M}\sqrt{1+sin(\theta_{\tt M 2})} \\
        R_{\tt I}\sqrt{1+sin(\theta_{\tt I 2})} \\
        1 \\
    \end{bmatrix} T_{\tt m} +
    \begin{bmatrix}
        0 \\
        0 \\
        r_{\tt M} \\
        - r_{\tt I} \\
        -1
    \end{bmatrix} T_{\tt s}(\bm{\theta}) 
    \label{Eq:actuation}
\end{eqnarray}
Here, $\tilde{\tau}_{\tt M 1}, \tilde{\tau}_{\tt I 1}, \tilde{\tau}_{\tt M 2}, \tilde{\tau}_{\tt I 2}, \tau_{\tt c}$
represent the actuation torque of MP joint and IP joint of the finger without the crawler,
MP joint and IP joint of the finger with the crawler, and the actuation force of the crawler, respectively.
$\theta_{\tt M 1}, \theta_{\tt I 1}, \theta_{\tt M 2}, \theta_{\tt I 2}$ are the joint displacement from the extended position
of those joints respectively. (See Fig.~\ref{fig:termination} (a).)
$T_{\tt m}$ is tendon tension due to the motor, which is the input to the hand model.
$T_{\tt s}(\bm{\theta})$ is the tendon tension in the sliding part,
which is a function of the hand state $\bm{\theta} = [\theta_{\tt M 1}, \theta_{\tt I 1}, \theta_{\tt M 2}, \theta_{\tt I 2}, \theta_{\tt c}]^T$.

The passive switching of the differential hand transmission is achieved by the switching of $T_{\tt s}$.
Its simulation model, though, is not straightforward due to the mutual dependency between the tendon tension and slider position,
i.e., the tendon tension $T_{\tt s}$ decides the slider position, while the slider position decides whether it reaches the end or not
thus affects $T_{\tt s}$.
To decouple them,
we consider an imaginal compliance $K$ (with a very high spring coefficient) as shown in Fig.~\ref{fig:termination} (b).
The displacement $x_{\tt t}$ is the supposed position of the slider derived from the hand state:

\begin{equation}
    x_{\tt t} = \theta_{\tt c} + r_{\tt I}\theta_{\tt I 2} - r_{\tt M}\theta_{\tt M 2} + x_{\tt ofs}
\end{equation}
Here, $\theta_{\tt c}$ is the displacement of the crawler and $x_{\tt ofs}$ is a constant offset.
The offset is to adjust the initial value of $x_{\tt t}$ to $0$, since the simulation starts with a non-zero
$\theta_*$ due to the parallel constraint.
The displacement $x_{\tt s}$ represents the actual position of the slider taking account of the motion limit.
If no motion limit is considered (denoted as $\tilde{x}_{\tt s}$), the following tension equilibrium stands:

\begin{align}
    K(x_{\tt t} - \tilde{x}_{\tt s}) &= k_{\tt s}\tilde{x}_{\tt s} + T_{\tt pt} \\
    \tilde{x}_{\tt s} &= \frac{Kx_{\tt t}-T_{\tt pt}}{K + k_{\tt s}}
\end{align}
Therefore, the slider position $x_{\tt s}$ can be expressed as:
\begin{align}
    x_{\tt s} = \begin{cases}
        0 & (\tilde{x}_{\tt s} < 0) \\
        \tilde{x}_{\tt s} & (otherwise) \\
        x_{\tt max} & (\tilde{x}_{\tt s} > x_{\tt max})
    \end{cases}
\end{align}
The three phases of $x_{\tt s}$ correspond to the three crawler driving modes described in the previous section.
Finally, the tendon tension $T_{\tt s}$ can be expressed as:
\begin{align}
    \tilde{T}_{\tt s} &= K(x_{\tt t} - x_{\tt s}) \\
    T_{\tt s} &= \begin{cases}
        \tilde{T}_{\tt s} & (\tilde{T}_{\tt s} > 0) \\
        0 & (otherwise) \\
        \end{cases}
\end{align}

The net joint torque is the summation of the extension spring torque and
the joint actuation torque described in Eq.~\ref{Eq:actuation}.
As shown in Fig.~\ref{fig:hand_schematic},
the IP joint extension spring connects the DP link with the parallel link,
rather than the PP link.
This is because that the latter case leads a contradiction between the extension springs of
the IP and MP joint.
In the latter case, when the hand is fully opened thanks to the MP extension spring,
the IP extension spring is maximally compressed,
thus working against the extension.

Inserting a torsional spring between the DP and parallel link results a coupling between the IP and MP joint.
We denote the coupling torque as $\tau_{p}$. ($p$ represents 'parallel'.)
Hereafter, we omit the subscript $\{1, 2\}$ for the two fingers because they have the same structure.
Since $\tau_{p}$ is due to the torsional spring, it can be expressed as:
\begin{align}
    \tau_{p} = \begin{cases}
        - k_e(\theta_{\tt I} - \tilde{\theta}_{\tt I} + \theta_{\tt e\_ofs}) & (\theta_{\tt I} > \tilde{\theta}_{\tt I}) \\
        K' (\theta_{\tt I} - \tilde{\theta}_{\tt I}) & (otherwise)
    \end{cases}
    \label{Eq:tau_p}
\end{align}
where $K'$ is a high gain to express the mechanical stopper, $\theta_{\tt e\_ofs}$ is the offset
of the torsional spring for the pre-load, $k_e$ is its spring coefficient,
and $\tilde{\theta}_{\tt I}$ is the parallel limitation
of the IP joint:
\begin{equation}
    \tilde{\theta}_{\tt I} = \theta_{\tt ofs} - \theta_{\tt M}
\label{Eq:theta_ofs}
\end{equation}
where $\theta_{\tt ofs}$ is the attach angle of the finger. (In this work, $2\pi/3$.)

The final joint torque sent to the physics engine can be written as:
\begin{align}
    \tau_{\tt I} &= \tilde{\tau}_{\tt I} + \tau_{p} - K'' \dot{\theta}_{\tt I} \\
    \tau_{\tt M} &= \tilde{\tau}_{\tt M} + \tau_{p} - k_e (\theta_{\tt M} + \theta_{\tt ofs}) - K'' \dot{\theta}_{\tt M} \label{Eq:tau_m}
\end{align}
where $K''$ is a damping gain to stabilize the simulation.
The second term of Eq.~\ref{Eq:tau_m} is because that $\tau_p$ contains $\theta_{\tt M}$,
as in Eq.~\ref{Eq:tau_p}, \ref{Eq:theta_ofs}.
It can be intuitively understood by considering the case
when $\theta_{\tt I}$ is fixed and $\theta_{\tt M}$ increases.
In that case the IP extension spring is stretched
since $\tilde{\theta}_{\tt I}$ changes according to $\theta_{\tt M}$ to maintain the parallel,
leading a extension torque on the MP joint.
The third term is due to its own extension torsional spring.

For the extension, we did not set the spring properties as optimized variables but fixed them as a constant 
with 0.1 Nm/rad stiffness and $\pi/6$ pretension.
In \cite{chen2019tendon}, the pulley radii were optimized for the post-contact torque and the extension springs
were optimized in the second layer for the pre-contact motion.
In our case, the pre-contact motion is regulated by the parallel link stopper mechanism therefore
the extension spring properties and the flexion pulley radii are redundant.
We set the former as constant to reduce the optimization complexity.

\subsection{Optimization}
In the simulation, the hand first vertically approaches the target object from the top until the fingertip
reaches the ground plane with a 1-mm margin.
The tendon tension then increases to the maximum value (in this case 100 N) with linear interpolation.
As the hand closes, the palm is lifted with feedback to keep the fingertip in a constant height.
After the hand is fully closed, the palm is lifted.
In the hand lifting stage, we apply a force-torque disturbance to the center of the object,
whose direction is random and magnitude is proportional to the lift height.
The disturbance is updated each 100 ms.
The maximum lift height (equivalent to the maximum magnitude of the disturbance)
before the object falls from the hand is recorded as the score.
Since the disturbance is random, we take an average of multiple trials.
Grasp of multiple different objects are simulated, and the scores are multiped to a single target value.
The overall optimization problem is shown as follows:
\begin{equation}
    \bm{\theta}_{\tt opt} = \argmax_{\bm{\theta}} \prod_{i=1}^n \left( 1+ \frac{1}{m}\sum_{j=1}^m h_j(p_i, \bm{\theta}) \right)
    \label{eq:opt}
\end{equation}
Here, $h_j(p_i, \bm{\theta})$ represents the maximum lift height of the $j$-th trial
of the object $p_i$, under the hand parameter $\bm{\theta}$.
$m$ is the iteration number to take the average, and $n$ is the number of objects.

To consider the score for multiple objects, we multiplied the score for each object as Eq.~\ref{eq:opt}.
We add an offset $1$ to each score to avoid grasp failure on a single object to lead
zero overall score, which increases the difficulty of the optimization problem.
To merge the scores, other choices were also possible.
One is to simply take the average of the scores 
to maximize $\frac{1}{n}\frac{1}{m}\sum_{i=1}^n\sum_{j=1}^m h_j(p_i, \bm{\theta})$.
In the optimization process, however, we found the optimizer only improved the score for easily graspable objects
such as those with a cylindrical shape and did not generate successful ones for thin and small ones.
We assume that it is because the weight of the improvement was constant through the height,
i.e., an improvement from 0 to 0.1 m height and the one from 1 to 1.1 m had the same weight.
Since improving an already graspable object by 0.1 m is easier than improving the hand to be able to
grasp an object that it could not pick before, the optimizer only focused on the former.
The formulation in Eq.~\ref{eq:opt} worked better since it is equivalent to
maximize the average of the log-scaled scores,
thus improvement from 0 to 0.1 m has a higher weight than the one from 1 to 1.1 m,
leading the optimizer to focus on difficult objects while also caring easy ones.
For the similar reason, to avoid that the optimizer only focuses on difficult objects and ignores easy ones,
we did not use the formulation to maximize the minimum score: $\min_i \frac{1}{m}\sum_{j=1}^m h_j(p_i, \bm{\theta})$,
since in this case a parameter update that improves graspable object does not affect the following score,
if there exists a not-yet graspable object.

In this work, we selected seven primitive shapes as the target objects ($n=7$),
namely a box with 50 mm $\times$ 10 mm width and thickness,
a box with 50 mm $\times$ 30 mm, a box with 150 mm $\times$ 10 mm,
a box with 150 mm $\times$ 30 mm, a cylinder with 8 mm diameter,
a cylinder with 20 mm, and a cylinder with 80 mm.
The depth, which is in the direction perpendicular to the plane in which the fingers move,
was set as 100 mm for all objects.
For each object and hand parameter, we simulated the grasp four times ($m=4$).

\begin{figure}[t]
    \begin{center}
        \includegraphics[width=1.0\columnwidth]{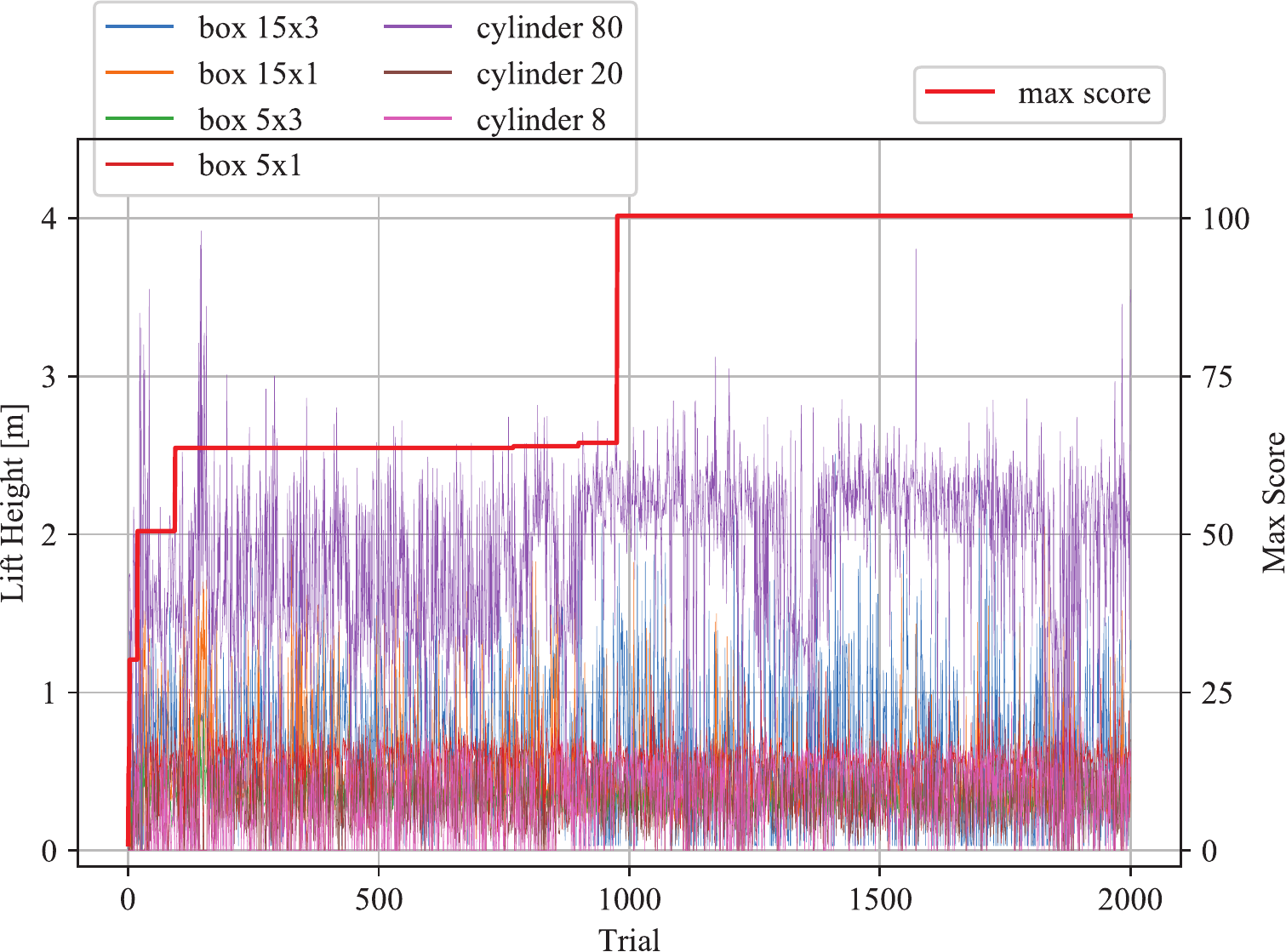}
        \caption{
            Transition of the object lift height and overall best grasp score against the iteration number.
        }
        \label{fig:score}
    \end{center}
\end{figure}

To solve the optimization problem, we used {\it Optuna}~\cite{optuna} as the framework.
While {\it Optuna} is originally developed for automatic hyperparameter search of machine learning projects,
we can also use it as a general-purpose black-box optimization tool.
As the optimization algorithm, we used the default Tree-structured Parzen Estimator (TPE)~\cite{tpe}.
For the physics simulation, we used PyBullet~\cite{pybullet}.
Figure~\ref{fig:score} shows the transition of the grasp score against the iteration number.
In this work, we ran the iteration for 2000 times, i.e., 56000 simulations in total.
The total computation time was 72 hours with a desktop PC with an 8-cored Intel Core i7 processor.
It reached the best score in the 976-th trial.

TABLE~\ref{tab:best_theta} shows the parameters for the best score.
The lower and upper bounds of each parameter are also shown.
The table shows that $R_{\tt M}$ saturates to the upper bound.
This is straightforward since a larger moment arm results in larger grasping force
and suggests that we can exclude this parameter from the optimization and
make it as large as space allows.
For $R_{\tt I}$, on the other hand, it is limited to a small value.
This is because that too large moment arm in the distal joint results in object ejection~\cite{underacuatedrobothands}.
Indeed, since $r_{\tt I}$ works in the antagonistic direction, the optimization shows that
the required effective moment arm of the IP joint is very small.
Object ejection is a common problem for underactuated hand, and in most of the hand design processes,
it needs to be explicitly considered.
In the proposed framework, on the other hand, the designer does not need to care about the problem since
the optimizer automatically avoids parameters that cause object ejection due to the low score.

\begin{table}[tb]
\caption{Design Parameters with the best Grasp Score}
\label{tab:best_theta}
\begin{center}
\begin{tabular}{cccc}
 & Optimal & L bound & U bound \\ \hline\hline
$l_{\tt D}$ & 74 mm & 40 & 80 \\ 
$l_{\tt P}$ & 92 mm & 60 & 120 \\ 
$l_{\tt M}$ & 80 mm & 40 & 80\\ 
$r_{\tt I}$ & 7 mm & 4 & $R_{\tt I}$-1\\ 
$R_{\tt I}$ & 8 mm & 8 & 12 \\ 
$R_{\tt M}$ & 20 mm & 10 & 20\\ 
$k_{\tt s}$ & 3.1 N/mm& 0.02 & 5\\ 
$T_{\tt pt}$ & 24 N& 0.1 & 50 \\ 
\end{tabular}
\end{center}
\end{table}

\begin{figure}[t]
    \begin{center}
        \includegraphics[width=1.0\columnwidth]{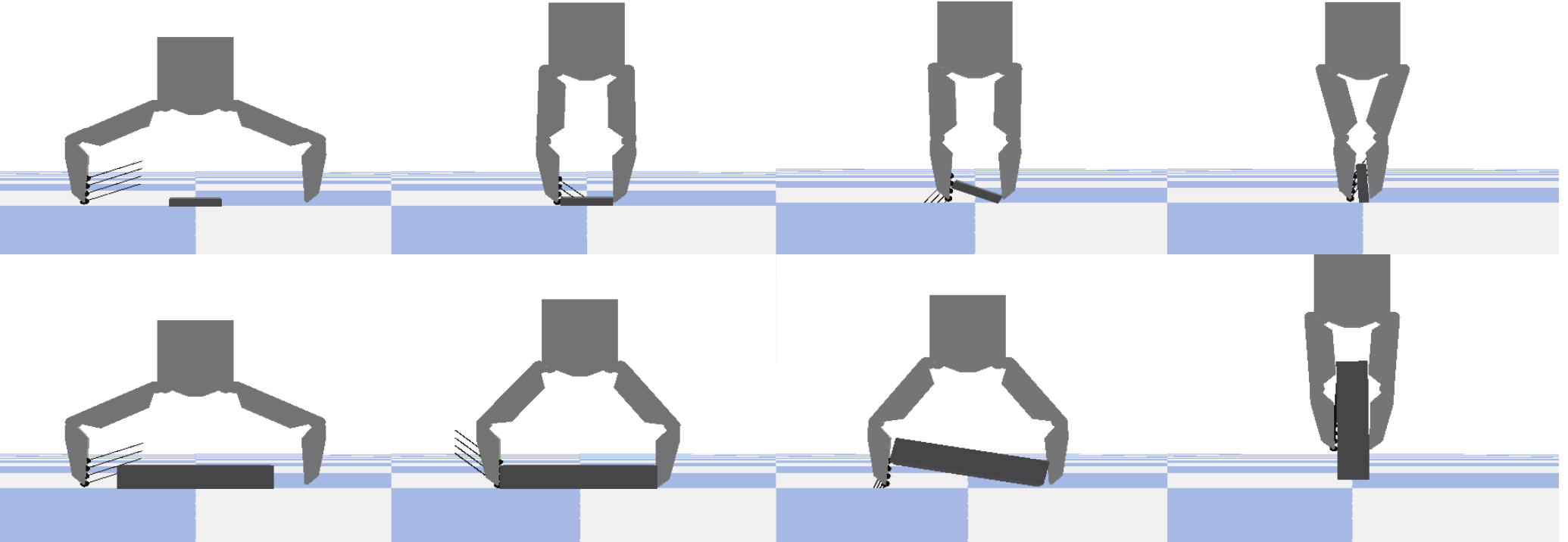}
        \caption{
            Transition of the grasp simulation
            of a 50 mm $\times$ 10 mm and 150 mm $\times$ 30 mm sized box
            when the parameters are the optimal one.
            The thin lines drawn from the fingertip crawler is to visualize the displacement of the crawler
            and not considered in the physics simulation.
        }
        \label{fig:sim}
    \end{center}
\end{figure}

In the simulation with the best score, two objects (box with 50 mm $\times$ 10 mm and
box with 150 mm $\times$ 30 mm) were grasped with the desired transition described in Fig.~\ref{fig:motion_schematic}.
Figure~\ref{fig:sim} shows rendered images of their time transition.
In those cases, the left side of the object is pulled up by the crawler.
When it reaches the end or the contact force exceeds the friction cone,
it rapidly slides to the upside.
Since the right side of the object is pushed toward the left side,
the object rotates clockwise to end up with a power grasp.
For the other two box-shaped objects, though, the grasp ended with a simple fingertip pinching.
(See the attached multimedia file for the result of all target objects.)
One reason is the modeling of fingertip crawler.
For simplicity, we modeled the crawler as a series of rollers.
In the grasps ended with a pinching grasp, the object's edge was trapped between the rollers
and not conveyed to the following one.
More realistic modeling of the crawler remains as our future work.

\section{Prototype Implementation and Evaluation}
\label{sec:evaluation}
Based on the parameters decided by the optimization, we prototyped a gripper shown in Fig.~\ref{fig:hand_outlook}.
The total weight is 750 g including a Dynamixel XM430 servomotor as the actuator and a custom-made automatic tool changer.
The whole parts except steel shafts and springs are 3D printed.
We used a high friction conveyer belt as the tendon/crawler.

\begin{figure}[t]
    \begin{center}
        \includegraphics[width=0.7\columnwidth]{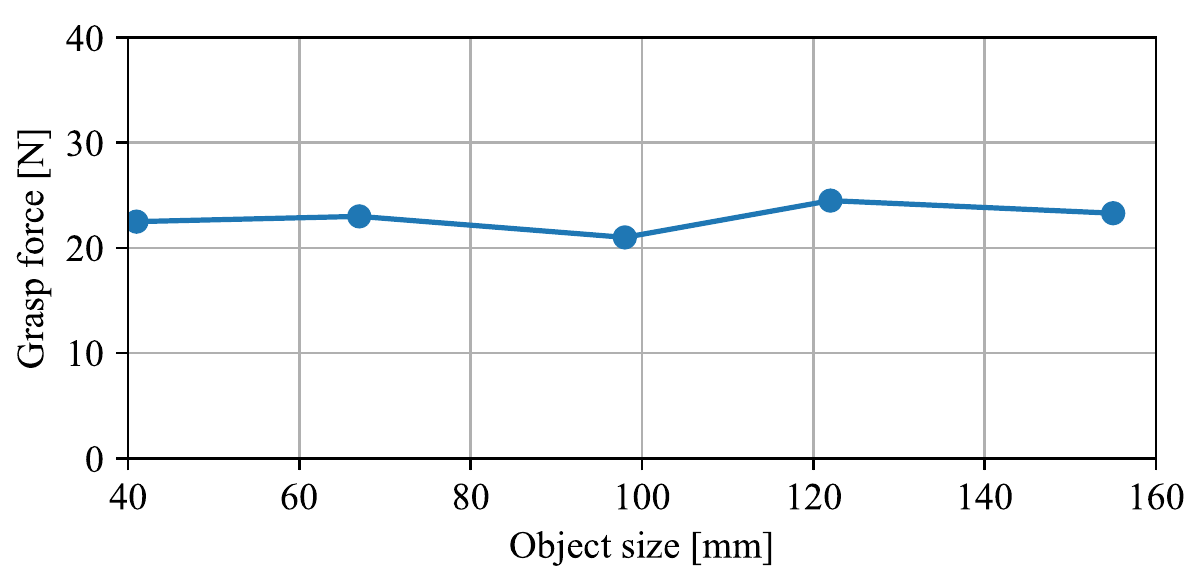}
        \caption{
            Measured maximum grasp force against object size.
        }
        \label{fig:grasp_force}
    \end{center}
\end{figure}

\begin{figure}[t]
    \begin{center}
        \includegraphics[width=0.9\columnwidth]{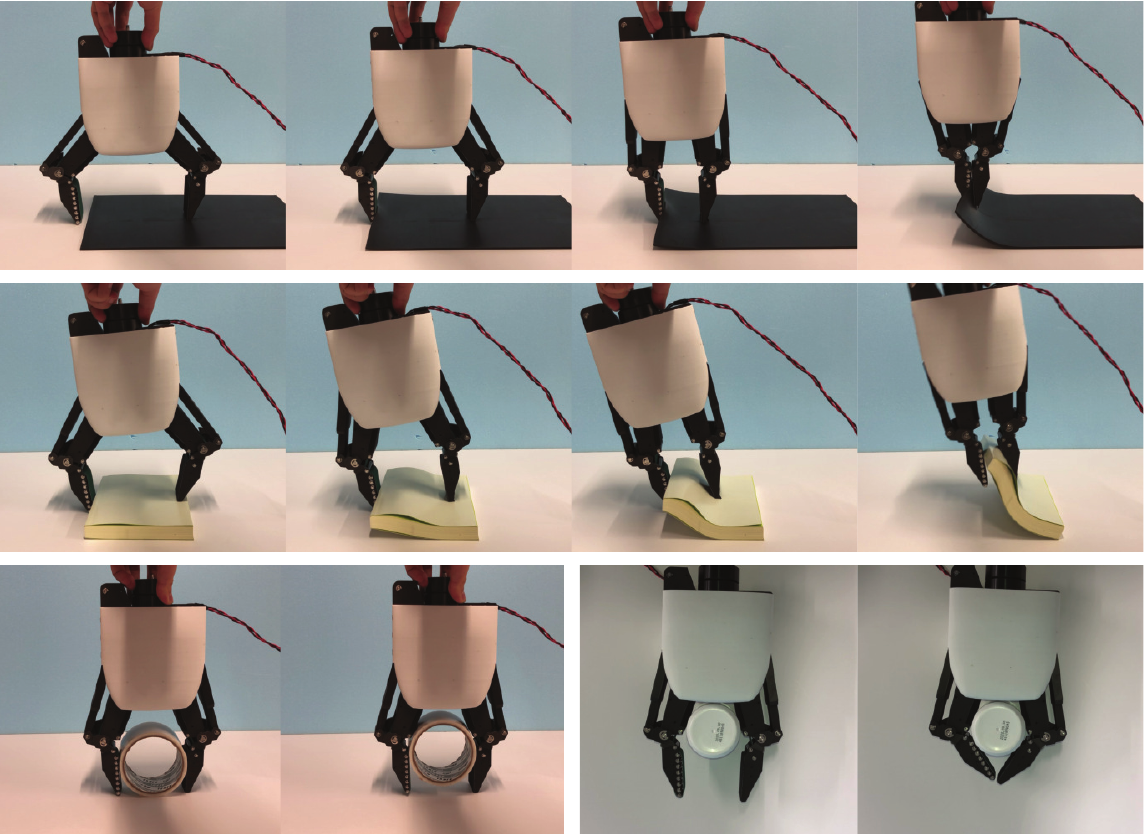}
        \caption{
            The proposed gripper can lift a 3 mm thick rubber sheet (top) and a softcover book (middle)
            lying on a flat surface.
            The fingertip crawler can lift the cylinder lying on the surface to end up with a power grasp. (bottom left)
            It can also envelop grasp small-sized objects. (bottom right)
        }
        \label{fig:video}
    \end{center}
\end{figure}

Figure~\ref{fig:grasp_force} shows measured maximum grasp force under parallel grasp configuration against multiple object sizes.
We attached spacers with multiple thicknesses to a force sensor and commanded the motor to exert a current equivalent to
100 N tendon tension to grasp the sensor.
The grasping force is flat against the object size and exceeds 20 N.
Figure~\ref{fig:video} shows demonstrations to lift flat and cylindrical objects.
On the top row, the gripper successfully picked a 3-mm-thick rubber sheet.
On the middle row, the gripper lifted a softcover book.
On the bottom left, the crawler lifted the cylinder lying on the surface to end up with a power grasp.
The bottom right demonstrates enveloping grasp ability.

\begin{figure}[t]
    \begin{center}
        \includegraphics[width=0.9\columnwidth]{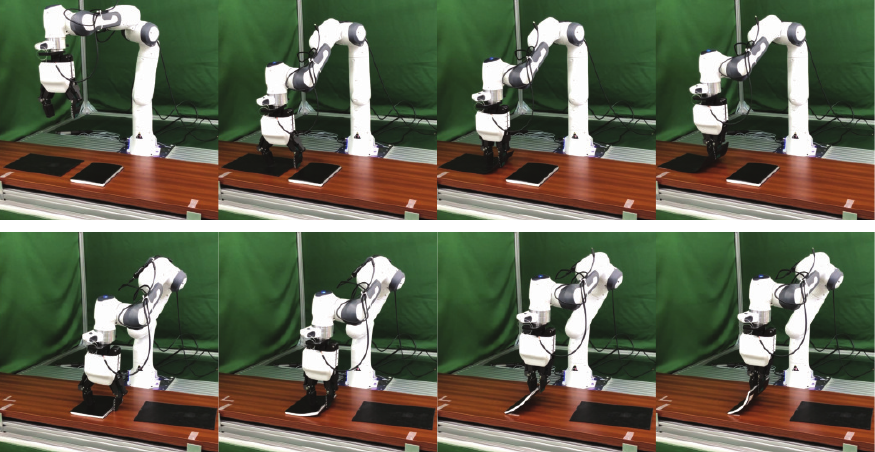}
        \caption{
            The gripper can lift thin objects with only a position-controlled arm without explicit force control.
            A rubber sheet is picked by simply command the arm to the pre-grasp position,
            close the hand with the maximum force, and lift the hand.
            For a hardcover notebook, an intermediate point with half grasp force and half lift was required.
        }
        \label{fig:panda}
    \end{center}
\end{figure}

In the previous demonstration, the hand was held by the human hand.
We found that proper manual control
utilizing the contact state and reaction force makes the grasping easy and stable.
To evaluate the grasp case without a sophisticated human control, we attached the hand to a Franka Emika Panda arm
and tried to grasp two objects: one is the 3 mm thick rubber sheet and the other is a hardcover notebook.
As the most straightforward control strategy, the arm was position-controlled under a commanded target cartesian point.
The rubber sheet was easily picked by simply command the arm to the pre-grasp position, close the hand with the maximum force,
and lift the hand.
The pre-grasp position was handcrafted.
For the notebook, though, the same strategy resulted in a slip between the crawler and the back of the notebook.
This is because the friction between them could not support the bending force of the notebook.
We therefore added another waypoint: the hand reaches to the pre-grasp position, close the hand with a half force,
lift by 10 cm, then grasp with the maximum force.
Figure~\ref{fig:panda} shows the view of the experiment.
See the attached multimedia file for the whole video, including the demonstration with the case held by the human hand.

The experiments show that the gripper can grasp thin objects with only a position-controlled arm
without explicit force control.
The trajectory, however, needs to be carefully handcrafted according to the object's property and friction condition.
Automatic generation of the motion and a feedback control scheme are required to adopt the hand
for real applications and remain as our future work.

\begin{figure}[t]
    \begin{center}
        \includegraphics[width=0.9\columnwidth]{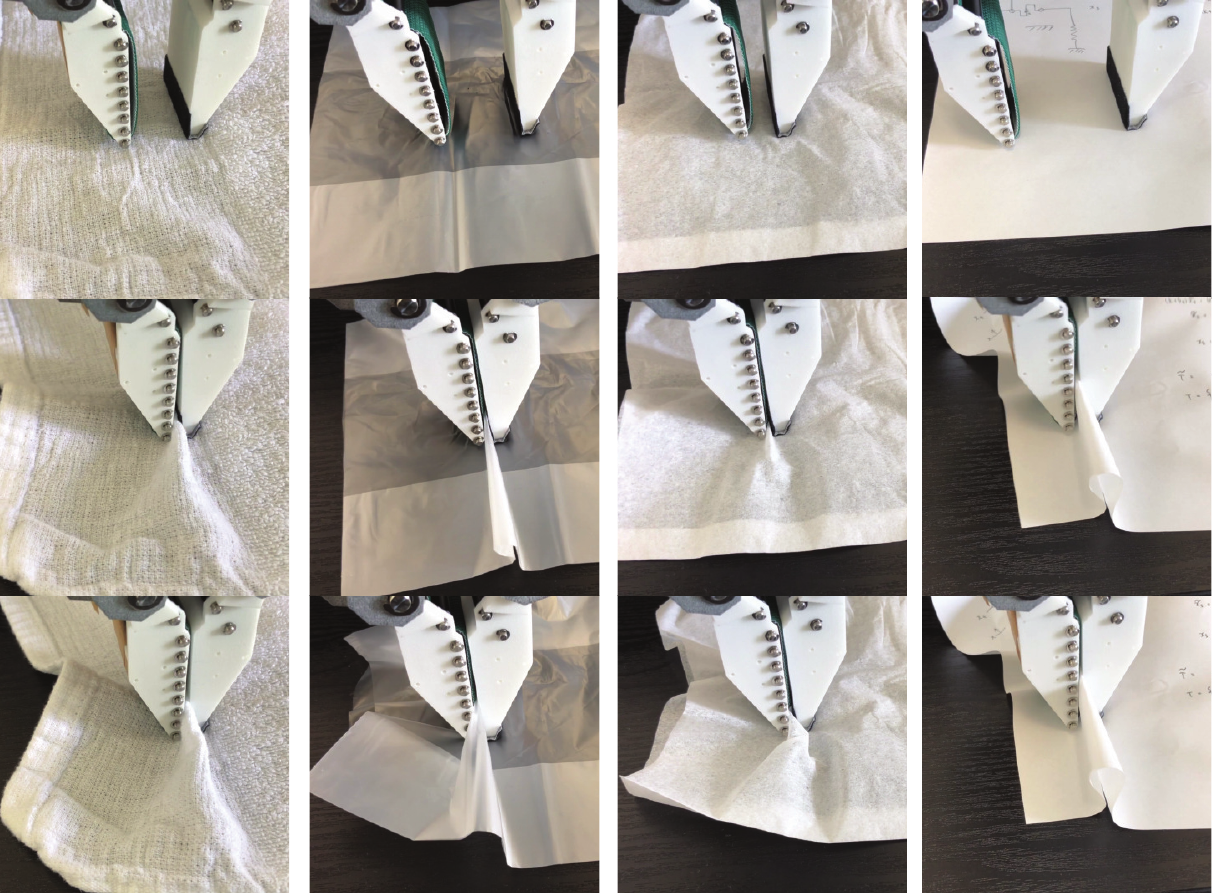}
        \caption{
            Pinching of a cloth (first column), a plastic bag (second column), a napkin (third column),
            and a sheet of paper (last column) from the top.
            The hand is hand-held as the same as the first experiment.
            The top row represents the initial state.
            The middle row in the state before the crawler moves.
            The last row is the last state after the crawler fully retracted.
            (See the attached multimedia file.)
        }
        \label{fig:pinching}
    \end{center}
\end{figure}

While not considered in the simulation, the experiments shown in Fig.~\ref{fig:pinching}
proved that the hand is also effective to approach from the top side of sheet-like objects,
to pick objects such as clothes, plastic bags, napkins, and paper.
In the initial stage of the picking, the crawler has little contribution, since the crawling force
is parallel to the ground surface thus it is the same with conventional parallel grippers.
(The transition between the top row to middle row in the figure.)
Once some part of the sheet is pinched, though, the crawler effectively pulled the pinched part
deeper in the hand to result in a stabler grasp,
which is difficult for conventional parallel grippers.
(Transition between the middle and bottom row.)

\section{Discussions and Limitations}
\label{sec:discussion}

\subsection{Comparison with existing grippers}
Comparing the proposed gripper with the existing parallel grippers such as \cite{robotiq_envelop, robotis},
ours is comparable with them in enveloping grasp, and advantageous in that it can in-hand manipulate a precision-grasped
object to a power-grasped configuration, leading a stabler grasp.
Its drawback, on the other hand, is the lack of precision-grasp ability:
for applications that require precise placement, such as assembly tasks,
in-hand motion of the object is not preferred.
One solution is to increase the stiffness of the spring in the slider part,
leading no crawler motion in the middle grasp-force range (thus enabling precision grasps)
and only perform the in-hand manipulation in the high grasp-force range.
Another solution is to have an active clutch on the slider to switch between the precision-grasp mode
and the power-grasp mode.
While this requires an additional actuator, it still keeps the advantage of the system's simplicity,
compared with adding another motor.

Comparing the proposed crawler approach with the {\it insert with the nail} approach \cite{babin2019stable, babin2018picking},
ours is more suitable for soft or high friction objects, since the {\it nail} approach requires the object
stiffness to stand against the friction between the nail and the object.
Ours is less suitable, on the other hand, for slippy objects or those that are thinner than the 
crawler tip radius.
The two approaches, though, are not exclusive but rather complemental: since both of the approaches only require one finger to be special,
we can have both of them in a single two-fingered hand to increase the variety of pickable objects.

\subsection{Reality gap of the simulation}
One of the large sources of the reality gap between the simulator and real hardware is friction in the transmission.
In this work, we did not model the friction thus the optimal result for the simulation might not be optimal
for the real case.
Nevertheless, the prototype performed as the original intension thanks to the mechanical adaptability.
Another possible reason is that since a hand optimization is fundamentally optimizing the ratio of the parameters,
rather than their absolute values, the effect of friction was not large.
The exception is the cases where the absolute value is important, such as the case discussed in the
previous subsection where the slider spring needs to be moderately strong to enable
both of the precision-grasp and power-grasp in the motor torque range.
In such cases, a two-stage approach is possible:
a prototype is built based on the initial optimization without considering friction,
then a system identification is performed for the prototype,
finally the optimization is performed again based on the model with the identified friction.

Another major source of the reality gap is the contact condition, such as surface smoothness and softness.
Those parameters are difficult to identify since they vary according to the environments and workpieces.
Instead of identifying a single value, one solution is to use the domain randomization~\cite{tobin2017domain}
to search for the design parameters that are robust against a predefined range of the contact parameters.

\subsection{Computation time}
One major limitation of the proposed optimization framework is the computation time: 
the 72 hours of computation time is around two orders larger than the 45 minutes in \cite{chen2019tendon}.
Its proportion in the total design period, though, is small compared with other hardware-related processes.
In this work we took around one week to 3D print all parts of the prototype with a single 3D printer,
and the whole prototyping process including designing, manufacturing, and assembling took around one month.

One straightforward approach is to leverage large-scale parallel computing with cloud computation resources.
Each simulation with the same parameter suggested by the optimizer is independent with each other and thus can be executed in parallel.
The optimization tool itself also supports parallel execution of multiple trials.
Another approach is to reduce the optimization dimension by 
statistically analyzing the effect of each parameter on the grasp performance.
Most of the commercially available hands have a series of lineup with the same basic structure (here we call it {\it meta-structure})
but different scales to handle workpieces with a large variety of sizes.
Through the optimization result for the meta-case,
we can derive the effect of each parameter or the ratio between them on the hand performance.
For the rest of the hands belonging to the same meta-structure, we can omit the parameters with less effect
and combine the related ones to reduce the overall computation time for the whole lineup of hands.
The black-box optimization assumes the problem to be non-differential, thus we cannot use the gradient information
to understand the effect of each parameter.
However, since the optimization process recursively runs sampling and evaluation, the accumulated record
of the sampled parameter and grasp performance makes the statistic analysis possible.
An alternative is to initially use random sampling to do the analysis
and secondly run the optimization with the reduced dimension.

\section{Conclusion}
\label{sec:conclusion}
The conclusion of this paper is as follows:
\begin{enumerate}
    \item We proposed a gripper with an underactuated surface on the fingertip.
          With the spring-loaded passive switching mechanism, actuation of a single motor
          generates three grasp modes in series: approaching the object as a standard parallel gripper,
          pulling the object inside the hand with the actuated fingertip crawler, and
          power grasping the object as an underactuated gripper.
    \item To optimize the design parameters, we proposed a simple approach
          to simulate the whole grasp process, that is approaching, pulling with the crawler, and power grasping
          until the hand finally dropped the object, and employed a general-purpose black-box optimization tool to solve this problem.
          The process can effectively consider the pre-contact motion, in-hand manipulation, power grasp stability, and even failure mode,
          which is difficult for the static-equilibrium-analysis-based approaches.
    \item We experimentally showed that a prototyped gripper with the proposed structure and design parameters optimized
          under the proposed process successfully picked a 3-mm-thick thin sheet and a softcover book from a flat surface,
          and lift a cylindrical shaped object from the surface to end with enveloping grasp.
\end{enumerate}

\section*{Acknowledgement}
The author would like to thank Hitoshi Kusano, Shimpei Masuda, Naoki Wakisaka, Dr. Naoki Fukaya, Dr. Tomomichi Sugihara,
Dr. Guilherme Maeda, Dr. Kuniyuki Takahashi, and Dr. Koichi Nishiwaki
for their support and fruitful discussions.




\bibliographystyle{IEEEtran}
\bibliography{bibs.bib}

\end{document}